\documentclass[a4paper,conference]{IEEEtran}

\usepackage{cite}
\usepackage{amsmath,amssymb,amsfonts}
\usepackage{algorithm}
\usepackage{algorithmic}
\usepackage{graphicx}
\usepackage{textcomp}
\usepackage{xcolor}
\usepackage[acronym]{glossaries}
\usepackage{booktabs}
\usepackage{xspace}
\usepackage[caption=false,font=footnotesize]{subfig}
\usepackage{pgfplots}
\pgfplotsset{compat=1.18}

\usepackage[a4paper, total={184mm,239mm}]{geometry}

\newcommand{\tool}{SPARC\xspace}

\newacronym{bptt}{BPTT}{Backpropagation Through Time}
\newacronym{snn}{SNN}{Spiking Neural Network}
\newacronym{cil}{CIL}{Class Incremental Learning}
\newacronym{lr}{LR}{Latent Replay}
\newacronym{tess}{TESS}{Temporally and Spatially Local Learning Rule}

\begin{document}
\bstctlcite{IEEEexample:BSTcontrol}
\title{\tool: Task-Free Class-Incremental Learning for Deep Spiking Neural Networks under a Local Rule}

\author{\IEEEauthorblockN{Lorenzo Capuano, Alessio Caviglia, Filippo Marostica, Alessandro Savino, Stefano Di Carlo}
\IEEEauthorblockA{Control and Computer Engineering Department\\
Politecnico di Torino, Torino, Italy\\
Contact: stefano.dicarlo@polito.it}
}

\maketitle

\begin{abstract}
On-device continual learning in deep SNNs must address catastrophic forgetting while relying on neuromorphic-compatible local learning rules. We present \tool, a task-free class-incremental framework that combines one-bit latent replay with TESS local learning algorithm. \tool aligns per-task partitions of trainable channels and TESS's fixed feedback/readout matrix, protecting prior knowledge while retaining a single global classifier over all observed classes. Layer-local readouts further reduce adaptation to one trainable convolutional layer, allowing the remaining pretrained tail to be discarded. Across four benchmarks with a frozen out-of-distribution VGG-9 backbone, \tool's peak training memory is constant in the number of timesteps, where BPTT's grows linearly with them, and each update costs $12\times$ fewer synaptic operations than BPTT. The saving against BPTT-based methods is already significant on static images ($2.3\times$ at $T{=}6$) and decisive on the long event sequences that motivate event-based sensing ($15.5\times$ at $T{=}60$). Despite that lower cost, \tool achieves $1.8$--$3.3$~pp higher accuracy than plain \gls{lr}, showing the gain is not merely a memory-saving trade-off.
\end{abstract}


\section{Introduction}
\glsresetall
Resource-constrained edge devices increasingly need to adapt after deployment to specific data, changing environments, sensor upgrades, and distribution shifts, while remaining within hardware budgets originally sized for inference~\cite{11130214}.
\glspl{snn} suit always-on event sensing: sparse activity enables low-power neuromorphic acceleration~\cite{merolla2014truenorth,davies2018loihi,pei2019tianjic,carpegna2022spiker,pehle2022brainscales,carpegna2025spikerplus,kudithipudi2025scale,leone2025syntzulu,spikerll}, while stateful dynamics process asynchronous sensor outputs directly~\cite{gallego2022}. Yet current accelerators largely support frozen inference. In-field learning should avoid the \gls{bptt} computationally expensive global backward pass and stored activation history, and must deal with catastrophic forgetting when adapting to new data~\cite{French1999Catastrophic,antonov}.

\gls{cil}, targeted by this paper, is one of the most complex continual learning problem settings. In \gls{cil}, disjoint classes arrive sequentially, but task identity is unavailable at inference and all observed classes must compete in one output space. Existing methods address only parts of this problem. \gls{lr} based methods reduce training data-storage cost but still rely on \gls{bptt}~\cite{spikinglr,replay4ncl}; local learning rules reduce adaptation cost without preventing forgetting~\cite{antonov}; and structural isolation requires non-local allocation or an auxiliary task classifier~\cite{dsdsnn}. Therefore, approaches to support deep \gls{snn} continual learning combining local adaptation, knowledge retention, and task-free inference still  remain not fully mature.

We close this gap with \tool (Sparse Partitioned Allocation for Replay-based Continual learning), to our knowledge the first framework to combine \gls{lr} with a local learning rule for task-free \gls{cil} in a deep \gls{snn}. \tool performs  every on-device update using the \gls{tess} over a one-bit latent buffer, while \gls{bptt} is confined to offline pretraining of a frozen backbone. The core mechanism ports per-task subnetwork allocation~\cite{dsdsnn} onto the fixed feedback matrix \gls{tess} already maintains: partitioning that matrix by task requires no growth-and-prune passes or learned synaptic masks, and the same feedback/readout partition makes inference class-incremental and task-free. Since \gls{tess} emits a prediction at every layer, the adapted tail can be reduced to a single trainable layer and the tail of the pretrained network dropped. Finally, we study backbone choice for realistic \gls{cil}, showing that out-of-domain backbone pretraining can suffice, but task-aligned pretraining is sometimes crucial.

Concretely, this paper presents: (i) a training-regime split that keeps global \gls{bptt} offline and confines on-device adaptation to a local rule, with the gain quantified as a measured peak-training-memory constant in $T$, $\approx15.5\times$ below \gls{bptt} at the same deployment cut; (ii) per-task subnetwork allocation realized entirely with the local rule's own operations, and made class-incremental and task-free by partitioning its feedback matrix; (iii) truncation of the backbone network, training of the last layer only, and discarding the remaining tail at low accuracy cost; (iv) the study of class-incremental \gls{snn} on an out-of-distribution frozen backbone; (v) an evaluation on different class-incremental benchmarks with a deep VGG-9 spiking backbone (most \gls{snn} continual-learning work adapts far shallower networks) reporting accuracy, backward transfer, and memory together, on which \tool achieves $1.8$--$3.3$~pp higher accuracy than plain latent replay despite its lower complexity and resource usage, showing that the gain is not merely a memory-saving trade-off.

\section{Background and Related Work}
\label{sec:background}

\subsection{Technical Background}
\label{subsec:snn-local-training}

Deep \glspl{snn} are commonly trained with surrogate gradients and \gls{bptt}. Unrolling the network stores state across all $T$ timesteps and requires a global backward path, whereas local rules maintain online eligibility traces and deliver learning signals directly to each layer~\cite{neurotrain}. Among several local rules proposed in the literature~\cite{neurotrain}, we build on \gls{tess}~\cite{tess}, which combines such traces with a fixed feedback matrix that both reads local class logits and projects their error back to the layer. Section~\ref{sec:methodology} defines the resulting update used by \tool.

\label{subsec:cl-background}
In \gls{cil}, disjoint classes arrive in successive training stages, but the task identity is unavailable at inference; the model must therefore rank all classes seen so far in one output space. Replay limits forgetting by mixing stored past samples with current data. \gls{lr} instead stores intermediate activations at a cut in a fixed feature extractor, reducing storage and avoiding recomputation of the preceding layers~\cite{ravaglia}; in an \gls{snn}, these latents can be one-bit spike tensors~\cite{spikinglr}.

\subsection{Related Work}

SpikingLR~\cite{spikinglr} and Replay4NCL~\cite{replay4ncl} compress \gls{lr} for shallow recurrent \glspl{snn}, but are based on \gls{bptt} for adaptation. Full-network rehearsal methods instead retrain all layers with replayed raw samples: STAER~\cite{staer} regularizes representations against forgetting, while Energy-Aware Spike Budgeting~\cite{energyaware} additionally controls spike rates. Both use a global backward path whose memory and data-movement costs \tool targets.

Structural methods isolate knowledge in task-specific capacity. DSD-SNN~\cite{dsdsnn} grows, prunes, and freezes subnetworks, with later work refining allocation~\cite{scasnn}; its class-incremental variant uses a separately trained task classifier at inference. PS-SNN~\cite{pssnn} improves this  by adding a feature extractor per task. CATFormer~\cite{catformer} freezes a spiking transformer after the first task and adapts task-specific thresholds and a head, but uses \gls{bptt} during adaptation and two backbone passes for task-free inference. These approaches trade additional training or model/inference complexity for isolation.

Local continual-learning rules avoid a global backward path but address a different point in the design space. HLOP~\cite{hlop} has been evaluated in task- and domain-incremental settings rather than \gls{cil}, while AGMP~\cite{agmp} updates a full ResNet-18 online without replay. Other nearby work either fits an analytic head using non-local operations, as in SLDA~\cite{slda,clpsnn}, or targets few-shot continual learning, as in SAFA-SNN~\cite{safasnn} and CLP-SNN~\cite{clpsnn}. None combines \gls{lr}, feature adaptation under a temporally and spatially local rule, structural protection against forgetting, and task-free class-incremental inference without an auxiliary task classifier.

\section{Methodology}
\label{sec:methodology}

\subsection{Framework Overview and Setting}
\label{subsec:framework-overview}

\begin{figure}[t]
    \centering
    \includegraphics[width=0.97\columnwidth]{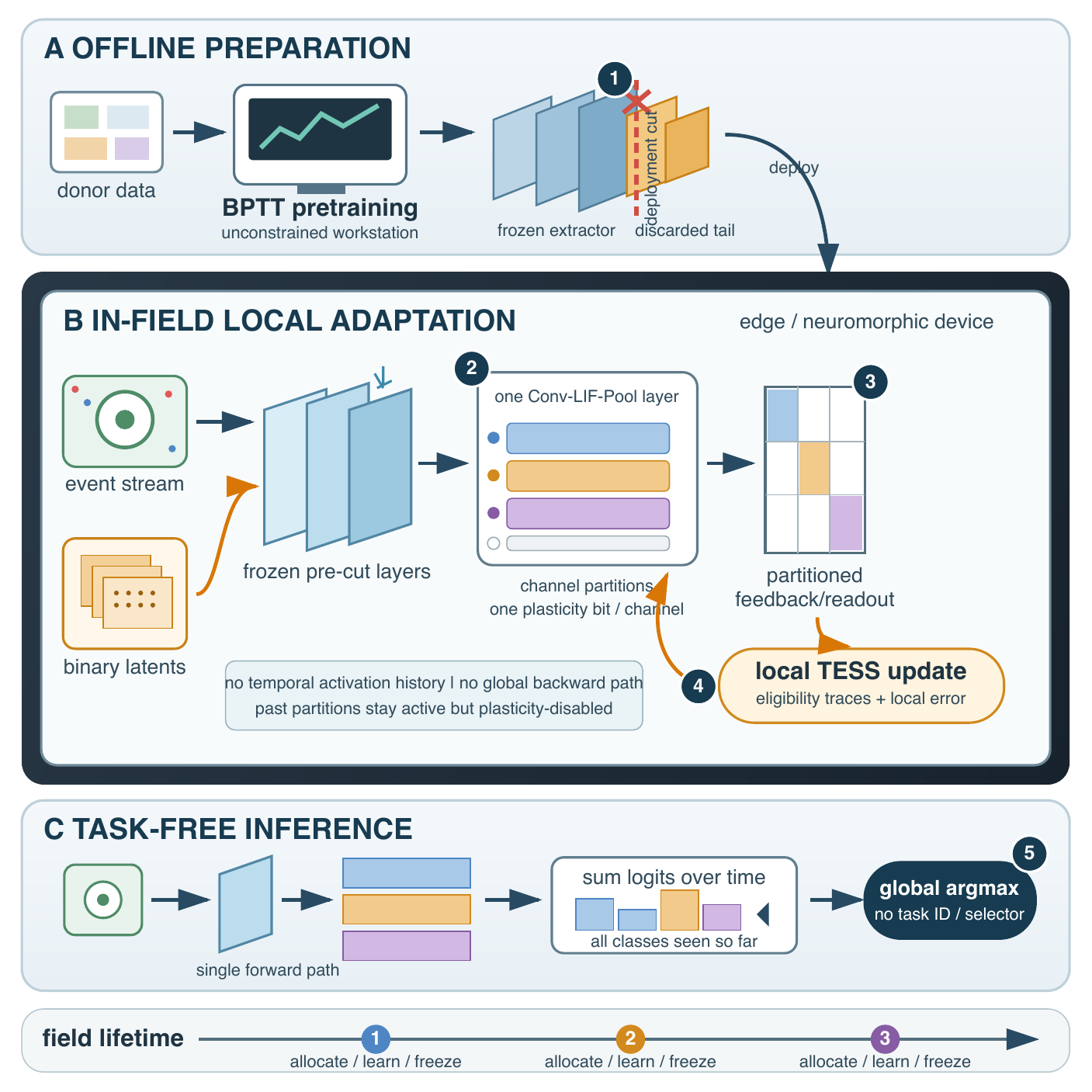}
    \caption{\tool deployment workflow. (A) Offline pretraining fixes the feature extractor at a deployment cut (1). (B) Current and replay latents feed one local layer; aligned channel (2) and feedback/readout (3) partitions gate TESS updates (4) while retaining prior partitions. (C) Accumulated logits support a task-free global argmax (5). The allocate--learn--freeze cycle repeats over the device lifetime.}
    \label{fig:sparc-framework}
\end{figure}

Figure~\ref{fig:sparc-framework} presents \tool as a complete deployment framework. Before deployment, an \gls{snn} backbone is pretrained offline, where a global rule such as \gls{bptt} is affordable. The network is then cut at a deployment-selected layer. We denote by $F_{\phi}$ the feature extractor formed by the layers before this cut; these layers are frozen, while the remaining tail is discarded and replaced by a single plastic spiking convolutional layer $G_W$. Once deployed, neither updates through the pre-cut layers nor a global backward path are required. The device learns each incoming class group by allocating free channels in $G_W$, coupling them to a new block of a fixed local feedback matrix, and enabling plasticity only in that partition. A binary latent buffer calibrates new and old class logits. At inference, all partitions are read jointly, so no task selector is needed.

We consider a sequence of $K$ training stages. Stage $k$ provides a dataset $\mathcal D_k$ with label set $\mathcal Y_k$, where $\mathcal Y_j\cap\mathcal Y_k=\emptyset$ for $j\neq k$. Stage boundaries and the labels in the current stage are available while learning, as in standard \gls{cil} protocols, while task identity is unavailable at inference. The device is configured with maximum class capacity $N_{\max}$ and channel capacity $C_{\max}$, but future task identities and samples are unknown. This distinction makes \tool boundary-aware during adaptation and task-free during operation.

For an input sequence $x_{1:T}$, the deployed forward path is
\begin{equation}
        z_t = F_{\phi}(x_t), \qquad
    s_t = G_W(z_t), \qquad
    h_t = P(s_t),
    \label{eq:sparc-forward}
\end{equation}
where $z_t$ is a binary latent, $s_t$ contains the adapter's output spikes, and $P$ is a fixed pooling and vectorization transform. $P$ converts the convolutional activity into the vector $h_t$ consumed by the TESS readout. Each channel contributes $p^2$ contiguous entries to $h_t$, where p is its pooling grid size.

\subsection{Frozen Backbone and Single-Layer Adapter}
\label{subsec:network_architecture_and_truncation}

The split in \eqref{eq:sparc-forward} separates representation extraction from in-field plasticity. Since the layers composing $F_{\phi}$ are fixed, current samples can be encoded as they arrive and old samples can be replayed directly from stored $z_{1:T}$, without retaining raw inputs or recomputing the pre-cut layers. The adapter $G_W$ contains exactly one Conv--LIF--Pool block. This is sufficient because \gls{tess} supplies a local readout and local learning signal at that layer; neither the discarded pretrained tail nor a conventional linear classifier participates in adaptation or inference.
This architecture is especially suitable for neuromorphic deployment. The fixed pre-cut layers follow the ordinary event-driven inference path, while the adapter needs only its current neuronal state, local eligibility traces, and a plasticity-enable bit per output channel. These quantities can remain local to the layer. In contrast to \gls{bptt}, no activation history over $T$ and no transport of errors through the layers before the cut are required.

\subsection{Channel Allocation and Partitioned Feedback}
\label{subsec:sparc_mechanism}

TESS attaches a fixed random matrix to each locally trained layer. On the forward path, this matrix maps the pooled layer activity $h_t$ to one logit per class; the same matrix maps the resulting class error back to that layer, where temporal eligibility traces turn it into a weight update. It therefore acts as both a local classifier and a feedback path, without propagating errors through later layers. \tool retains this TESS mechanism but partitions the matrix and the adapter channels together at each stage.
Let $\mathcal F_k\subseteq\{1,\ldots,C_{\max}\}$ be the channels still free before stage $k$. The number assigned to the new labels is:
\begin{equation}
    n_k =
    \left\lfloor
    \gamma_k
    \left\lfloor\frac{C_{\max}}{N_{\max}}\right\rfloor
    |\mathcal Y_k|
    \right\rfloor,
    \label{eq:channel-allocation-new}
\end{equation}
where $\gamma_k\in(0,1]$ controls the fraction of the configured per-class capacity used at stage $k$. \tool selects a set $\mathcal A_k\subseteq\mathcal F_k$ with $|\mathcal A_k|=n_k$. Allocated channels are never reassigned. Their channel-level plasticity vector is:
\begin{equation}
    a_{k,c}=\mathbb{I}[c\in\mathcal A_k].
    \label{eq:plasticity-gate}
\end{equation}
where $\mathbb{I}[\cdot]$ is the indicator function. Thus, $a_k\in\{0,1\}^{C_{\max}}$ has one entry per adapter channel: entries for $\mathcal A_k$ enable learning, whereas all others suppress updates without removing those channels from the forward path. Unlike importance-based growth and pruning, this gate is written once at a stage boundary and requires no global sensitivity estimate.

Formally, let $d$ be the dimension of $h_t$ and $B^{(k)}\in\mathbb R^{N_{\max}\times d}$ the feedback/readout matrix after stage $k$. Each row of $B^{(k)}$ represents a class, while each column receives one coordinate of $h_t$. If $P$ produces a $p\times p$ grid per channel, every channel contributes $p^2$ coordinates. The columns associated with the new channels are:
\begin{equation}
    \mathcal J_k=
    \{(c,r,q):c\in\mathcal A_k,\ 0\leq r,q<p\}.
    \label{eq:feedback-columns}
\end{equation}
Each tuple $(c,r,q)$ is flattened to one column index $j$, so $\mathcal J_k$ contains $p^2$ columns per allocated channel. Starting from $B^{(0)}=0$, \tool writes fixed random TESS values $R_k$ only where the current-class rows intersect these current-channel columns:
\begin{equation}
 B^{(k)}_{y,j}=
 \begin{cases}
   (R_k)_{y,j}, & y\in\mathcal Y_k,\ j\in\mathcal J_k,\\
   B^{(k-1)}_{y,j}, & \text{otherwise}.
 \end{cases}
 \label{eq:feedback-block}
\end{equation}
The resulting matrix is block sparse: current classes read only the current channel partition, and their errors are returned only to that partition. Earlier blocks remain fixed, while cross-stage entries remain zero. Together, $a_k$ and $B^{(k)}$ impose the same partition on learning and readout. The physical matrix may be preallocated or populated incrementally; only this support pattern matters.

At timestep $t$, the local logits, class error, and neuron-local modulatory signal are:
\begin{align}
    \ell_t &= B^{(k)}h_t, &
    e_t &= \operatorname{softmax}(\ell_t)-\mathbf y, \label{eq:local-error}\\
    g_t &= P^{\top}(B^{(k)})^{\top}e_t, && \label{eq:local-modulation}
\end{align}
where $\mathbf y$ is the one-hot target. The first line is the readout role of $B^{(k)}$: it compares all class logits in one space. The second is its feedback role: $(B^{(k)})^{\top}$ returns the class error to the corresponding readout coordinates, and $P^{\top}$ distributes it to the adapter neurons. No error traverses later network layers. Aligning these two roles with the channel gate is the central \tool mechanism.

\subsection{Locally Gated Adaptation}
\label{subsec:local-adaptation}

The feedback signal above solves spatial credit assignment locally; TESS handles temporal credit assignment with running traces rather than a stored history. \tool does not alter these TESS recurrences: its addition is the channel gate applied to their final update. Let $q_t$ and $r_t$ denote decaying pre- and postsynaptic traces:
\begin{equation}
    q_t=\lambda_{\mathrm{pre}}q_{t-1}+z_t,
    \qquad
    r_t=\lambda_{\mathrm{post}}r_{t-1}+\Psi(u_{t-1}),
    \label{eq:tess-traces}
\end{equation}
where $u_t$ is the membrane potential and $\Psi$ is the surrogate response around the firing threshold. The causal and non-causal eligibility terms are:
\begin{equation}
    E_t^{\mathrm{pre}}=\Psi(u_t)\otimes q_t,
    \qquad
    E_t^{\mathrm{post}}=r_t\otimes z_t,
    \label{eq:tess-eligibility}
\end{equation}
with $\otimes$ interpreted as the local outer product, or its convolutional cross-correlation equivalent. Using the modulatory signal in \eqref{eq:local-modulation}, the update is
\begin{equation}
    \begin{aligned}
    \Delta W_t &= -\eta\,\operatorname{diag}(a_k)\bigl[g_t\odot{}\\
    &\quad \bigl(\alpha_{\mathrm{pre}}E_t^{\mathrm{pre}}+
          \alpha_{\mathrm{post}}E_t^{\mathrm{post}}\bigr)\bigr].
    \end{aligned}
    \label{eq:sparc-update}
\end{equation}
Broadcasting $a_k$ over each output filter enables updates only for the current channels. Past filters still take part in the forward pass, but their plasticity bits are zero. This is a coarse channel gate, not a learned or unstructured synaptic mask; it can be implemented by suppressing local update events for disabled channels. Equations~\eqref{eq:tess-traces}--\eqref{eq:sparc-update} are evaluated forward in time, so their state is independent of sequence length.

\subsection{Binary Latent Replay and Stage Algorithm}
\label{subsec:latent_replay_setup}

Structural isolation protects past filters but does not by itself calibrate logits produced by different blocks. \tool therefore mixes current latents with replay from previously observed classes. For a minibatch $\mathcal B_k=\mathcal B_{\mathrm{curr}}\cup\mathcal B_{\mathrm{replay}}$, the target replay fraction is:
\begin{equation}
    \rho_k=
    \frac{N_{\mathrm{past}}}{N_{\mathrm{past}}+N_{\mathrm{curr}}},
    \qquad
    |\mathcal B_{\mathrm{replay}}|=\lfloor\rho_k|\mathcal B_k|\rfloor.
    \label{eq:replay-ratio-new}
\end{equation}
For an old-class replay sample, the old block already supplies its frozen positive logit, while the local error suppresses competing logits produced by the new block. Replay therefore trains cross-task comparability rather than relearning old filters.

At the end of stage $k$, a uniformly sampled fraction of $\mathcal D_k$ is encoded once and appended to the persistent buffer:
\begin{equation}
    \mathcal R_k=\mathcal R_{k-1}\cup
    \operatorname{Sample}\!\left(F_{\phi}(\mathcal D_k),m_k\right).
    \label{eq:buffer-update}
\end{equation}
The stored values are binary spike tensors and can therefore be bit-packed in a deployed implementation. The buffer is needed only during later adaptation, not during inference.

Algorithm~\ref{alg:sparc} summarizes the complete adaptation procedure for class-incremental stage $k$, from channel allocation through local training and buffer update.

\begin{algorithm}[t]
\footnotesize
\caption{\tool adaptation at class-incremental stage $k$}
\label{alg:sparc}
\begin{algorithmic}[1]
\REQUIRE $\mathcal D_k$, fixed pre-cut extractor $F_{\phi}$, adapter $W$, feedback $B^{(k-1)}$, free channels $\mathcal F_k$, buffer $\mathcal R_{k-1}$
\STATE Determine current label set $\mathcal Y_k$
\STATE Allocate $\mathcal A_k\subseteq\mathcal F_k$ using \eqref{eq:channel-allocation-new}
\STATE Set plasticity vector $a_k$ using \eqref{eq:plasticity-gate}
\STATE Write feedback block $\mathcal Y_k\times\mathcal J_k$ using \eqref{eq:feedback-block}
\FOR{each adaptation minibatch}
    \STATE Mix current and replay samples according to \eqref{eq:replay-ratio-new}
    \STATE Encode current samples with $F_{\phi}$; read past latents from $\mathcal R_{k-1}$
    \FOR{$t=1,\ldots,T$}
        \STATE Update neuron states and eligibility traces
        \STATE Compute $\ell_t$, $e_t$, and $g_t$ using \eqref{eq:local-error}--\eqref{eq:local-modulation}
        \STATE Apply the locally gated update in \eqref{eq:sparc-update}
    \ENDFOR
\ENDFOR
\STATE Freeze $\mathcal A_k$ and remove it from the free-channel set
\STATE Update the binary latent buffer using \eqref{eq:buffer-update}
\end{algorithmic}
\end{algorithm}

\subsection{Task-Free Inference and Persistent State}
\label{subsec:task-free-inference}

After stage $k$, inference uses the same forward path for every input. Per-timestep logits are accumulated and ranked over all observed classes:
\begin{equation}
    \begin{aligned}
    L(x)&=\sum_{t=1}^{T}B^{(k)}P\!\left(G_W(F_{\phi}(x_t))\right),\\
    \hat y&=\arg\max_{y\in\mathcal Y_{1:k}}L_y(x).
    \end{aligned}
    \label{eq:task-free-inference}
\end{equation}
No task-conditioned routing occurs: the block that produces the largest class logit implicitly identifies both task and class. The persistent deployment state comprises the fixed layers before the cut, one adapted convolutional layer, the block-sparse feedback matrix, one allocation bit per channel, and the replay buffer used for future updates. No auxiliary task classifier, past activation history, or backward network is stored.

\section{Results}
\label{sec:results}

\begin{figure}[t]
\centering
\includegraphics[width=\columnwidth]{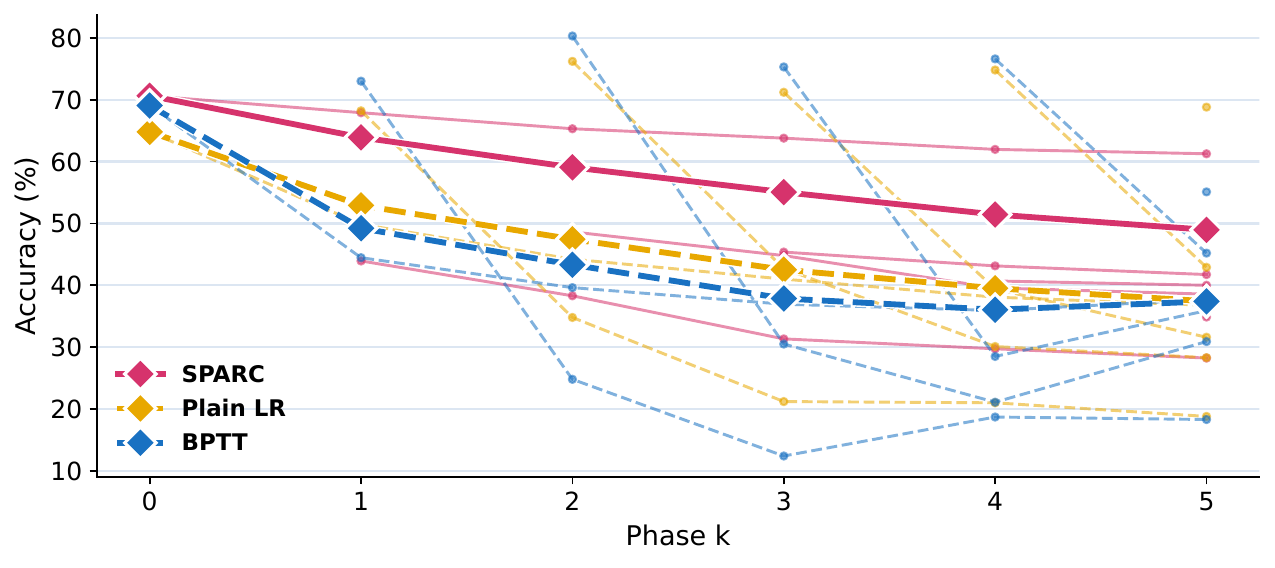}
\caption{Per-task (thin) and overall (bold) accuracy on CIFAR-100 B50-Inc10. Plain LR and \gls{bptt} learn new tasks better but forget them quickly; \tool retains them.}
\label{fig:c100-trajectory}
\end{figure}

\begin{figure*}[t]
    \centering
    \subfloat[Accuracy and BWT\label{fig:insertion-accuracy-bwt}]{
        \includegraphics[width=0.45\textwidth]{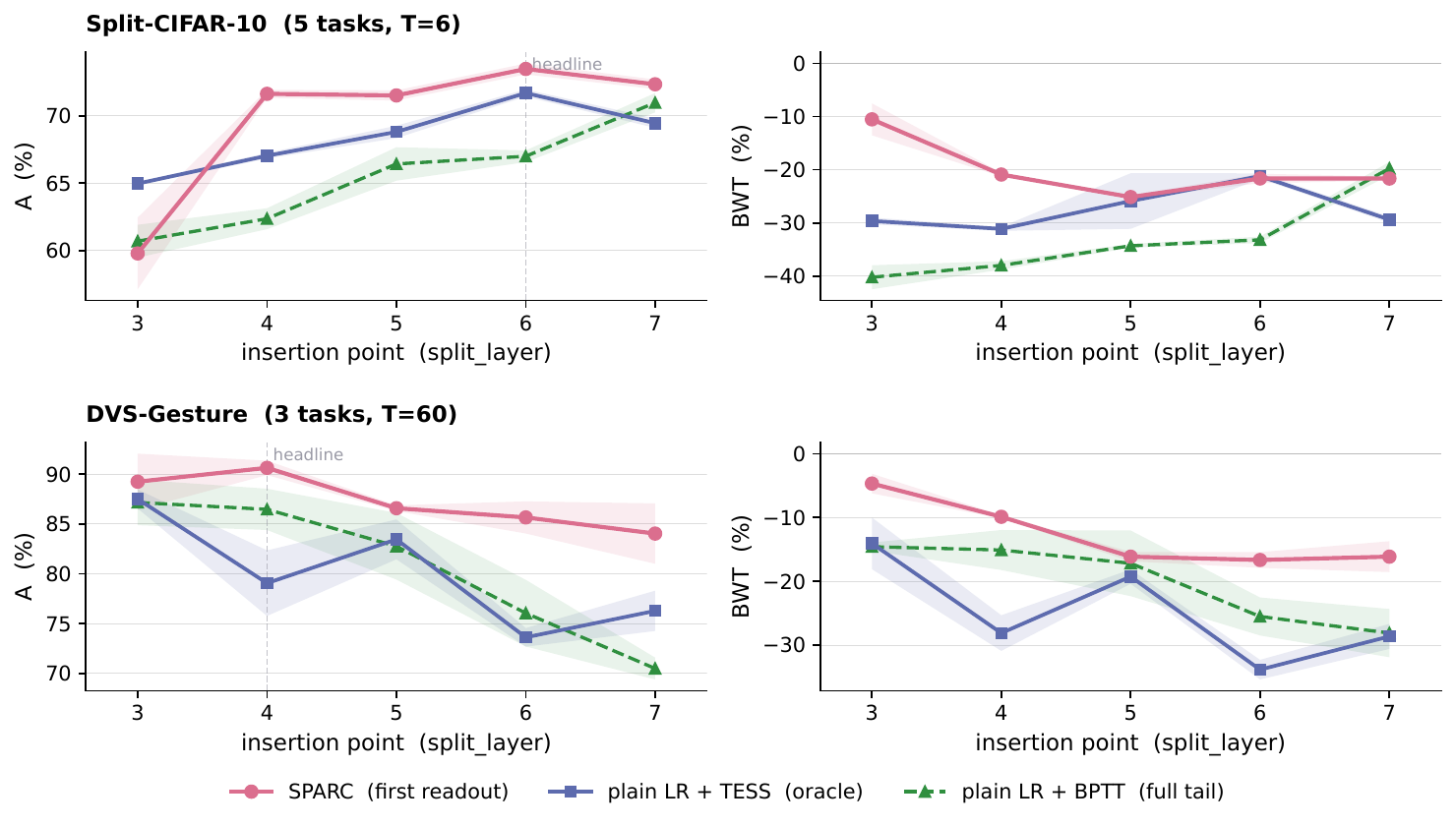}
    }
    \hfill
    \subfloat[Training memory and synaptic operations \label{fig:insertion-cost}]{
        \includegraphics[width=0.45\textwidth]{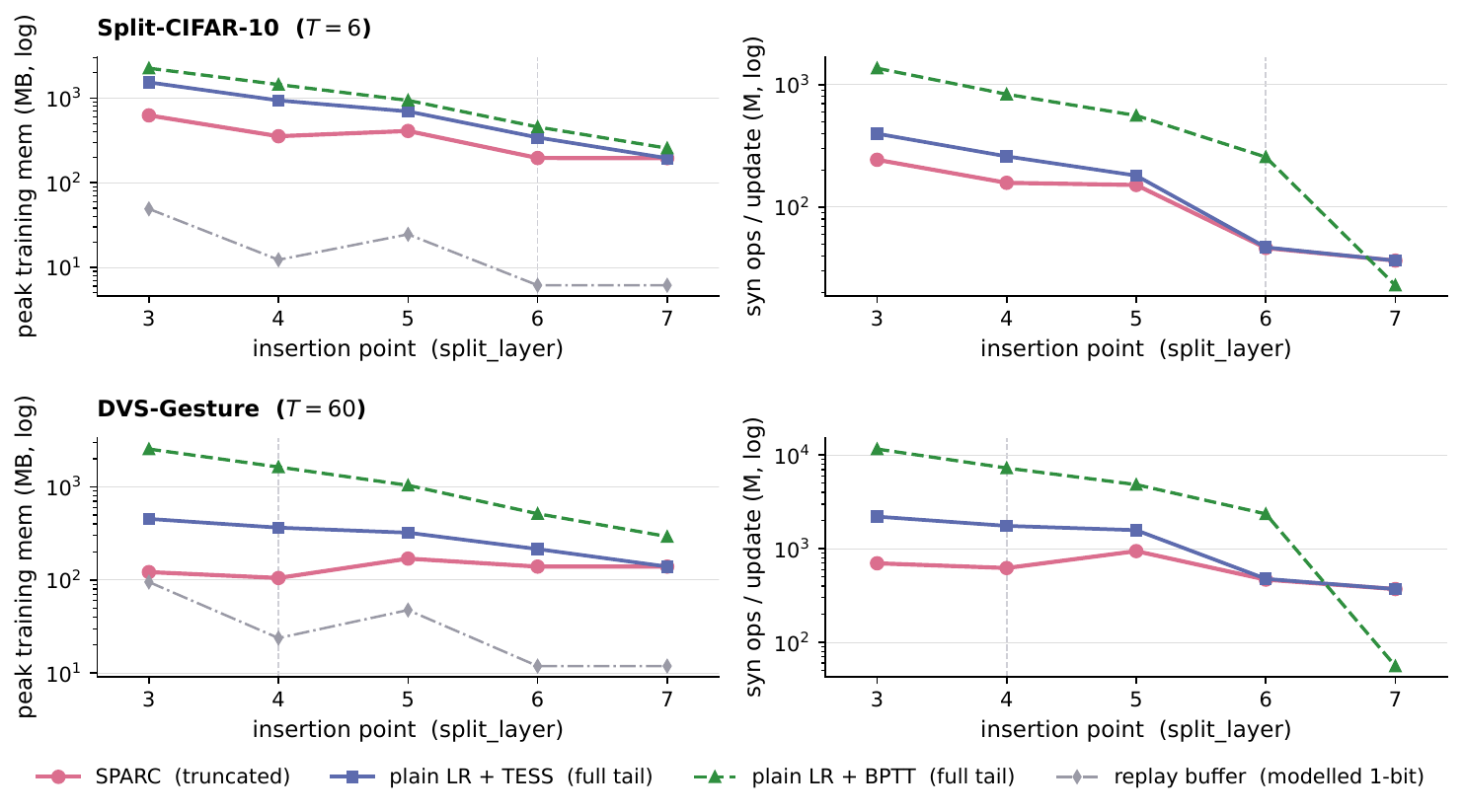}
    }
    \caption{Effect of the insertion point on Split-CIFAR-10 and DVS-Gesture. (a) Single-layer \tool matches or exceeds plain LR's best-case oracle readout at nearly every point. (b) Peak training memory (left) and measured synaptic operations (right) per update, log scale, at each dataset's own $T$ and batch size. The replay buffer is modelled at one bit per element.}
    \label{fig:combined_results}
\end{figure*}

\subsection{Experimental Setup}
\label{subsec:setup}
We focus on demonstrating the capability of the \tool approach while the deployment on a target embedded system is out of the scope of this paper. Models are implemented and simulated in snnTorch \cite{snntorch}.
We evaluate thoroughly with Split-CIFAR-10 \cite{cifar} ($5\times2$ classes, $T{=}6$) and DVS-Gesture \cite{dvsgesture} ($4{+}4{+}3$ classes, $T{=}60$). Additionally, we target complex datasets to show the limitations of the local training methods: Split-CIFAR-100 \cite{cifar} ($10\times10$ classes, $T{=}6$), and CIFAR10-DVS \cite{cifardvs} ($5\times2$ classes, $T{=}10$). $T$ is the number of simulation timesteps used to process each sample, splits are given as tasks~$\times$~classes per task, or as a sum of per-task class counts where these are uneven. A VGG-9 \cite{vgg} \gls{snn} backbone is pretrained by \gls{bptt} on disjoint data (CIFAR-100 for CIFAR-10; N-MNIST \cite{nmnist} for DVS-Gesture) and frozen through the insertion point before a one-layer spiking convolutional tail trained by \gls{tess}\footnote{Our \gls{tess} implementation uses per-layer random feedback matrices because, following the authors' advice, we could not reproduce the published square-wave construction.}. The one-bit latent buffer retains $2\%$ (CIFAR) or $5\%$ (DVS) of each task's training set. We report final average class-incremental accuracy $A$ and backward transfer BWT in percentage points, averaged over three fresh seeds; sample standard deviations are reported where comparisons depend on their spread. Plain \gls{lr} (dense feedback, no freezing) is given a per-run oracle over its layer readouts, providing a best-case baseline, except in Figure \ref{fig:ablation} where it matches \tool's single trainable layer.

\subsection{Comparison with Plain LR}
\label{subsec:comparison_lr}

\tool's design goal is to cut on-device learning cost; despite that saving, it matches or exceeds plain latent replay's own readout-oracle accuracy on every dataset, comparing each method at its own best-performing insertion point: $+1.8$~pp~$A$ / $-0.45$~pp~BWT on Split-CIFAR-10 ($73.5\pm0.5$ vs $71.7\pm1.1$; $-21.64\pm0.9$ vs $-21.19\pm1.3$), $+3.3$ / $+4.6$ on DVS-Gesture ($90.6\pm0.7$ vs $87.3\pm2.1$; $-9.9\pm0.2$ vs $-14.5\pm2.6$), and $+19.4$~pp~$A$ / $+46.9$~pp~BWT after ten Split-CIFAR-100 tasks ($36.3\pm0.5$ vs $16.9\pm0.3$; $-27.2\pm1.3$ vs $-74.1\pm0.4$).

\begin{figure}[t]
    \centering
    \subfloat[Buffer size sensitivity.\label{fig:buffer-accuracy-bwt}]{%
        \includegraphics[width=\columnwidth]{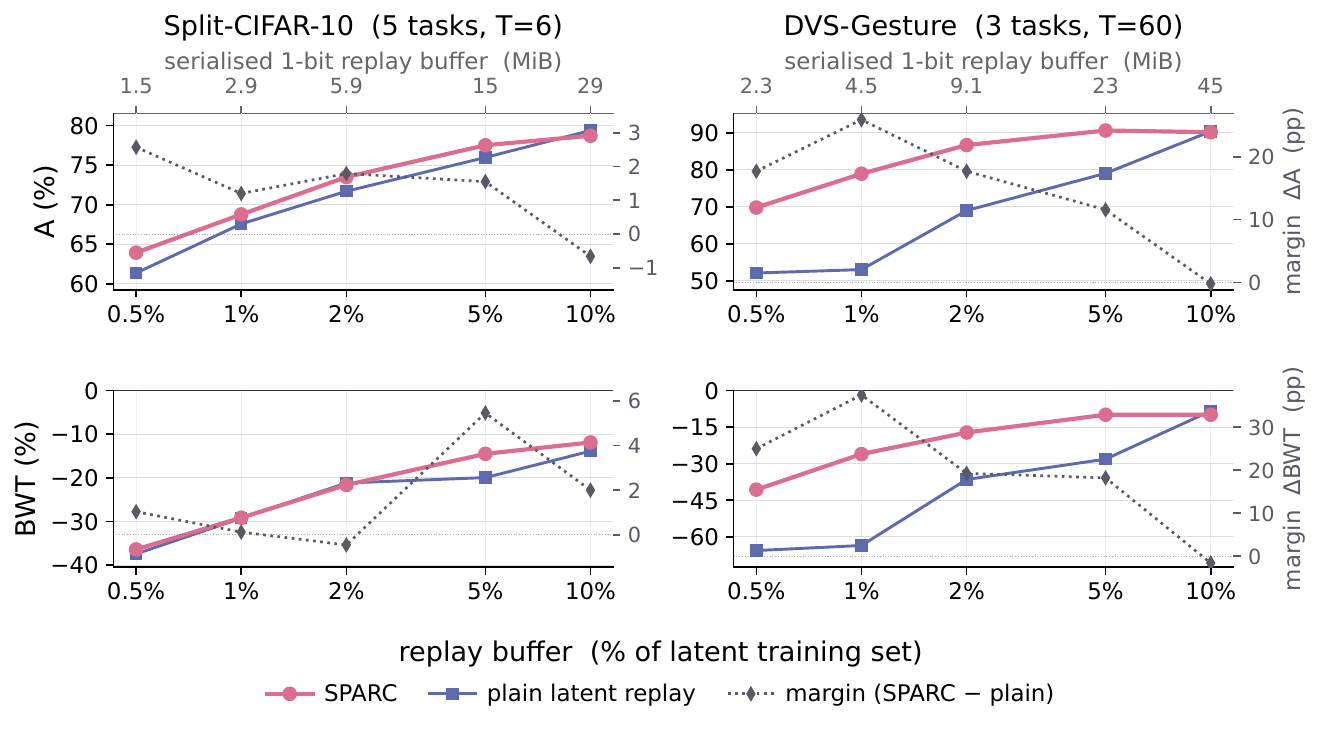}%
    }\\
    \subfloat[Pretraining donor sensitivity.\label{fig:donor-transfer}]{%
        \includegraphics[width=\columnwidth]{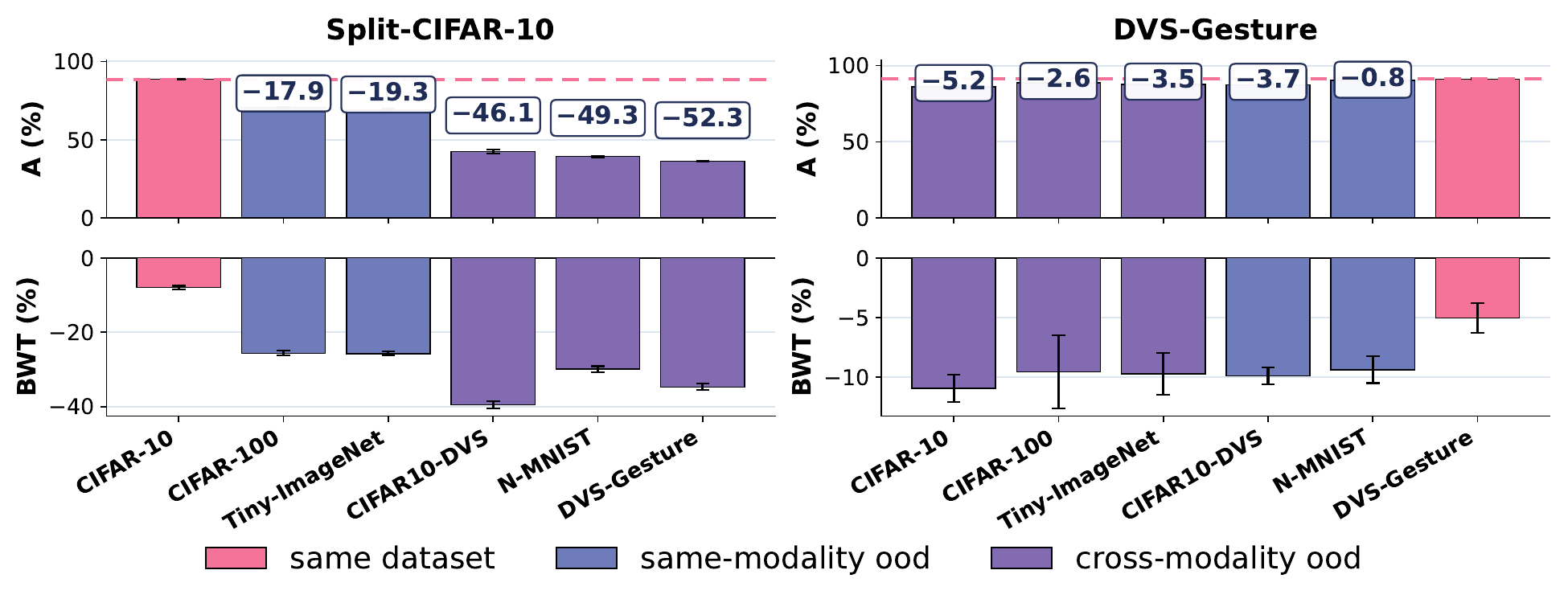}%
    }

    \caption{Sensitivity at the deployed insertion points.
    (a) Accuracy and BWT versus buffer size; \tool's margin grows as the
    buffer shrinks.
    (b) Final accuracy and BWT across pretraining donors, grouped as oracle
    (same dataset), same-modality out-of-distribution (OOD), or cross-modality OOD; labels show the oracle accuracy gap. Split-CIFAR-10 is sensitive even within
    modality, whereas DVS-Gesture is robust across modalities.}
    \label{fig:sensitivity}
\end{figure}

Three further studies broaden this comparison. First, across five VGG-9 insertion points (Fig.~\ref{fig:insertion-accuracy-bwt}), \tool matches or exceeds the plain \gls{lr} oracle readout beyond the earliest point tested on each dataset, with similar BWT.
The best point is later on Split-CIFAR-10 but earlier on DVS-Gesture, consistent with their different layer-wise peaks. Second, \tool leads throughout the buffer sweep (Fig.~\ref{fig:buffer-accuracy-bwt}), giving better accuracy at every fixed buffer size and its largest margin at the smallest buffers.
Third, in a large-scale CIFAR-100 setting with a 50-class base task followed by five 10-class increments, shown in Fig.~\ref{fig:c100-trajectory}, \tool forgets $7.9$~pp against $57.8$~pp for plain \gls{lr} and $52.3$~pp for \gls{bptt}, reaching final accuracy of $49.0\%$ versus $37.5\%$ and $37.4\%$. Reduced plasticity thus lowers new-task learning relative to plain \gls{lr} and \gls{bptt}, but the aligned block-sparse feedback compensates for it through retention.

This margin requires no larger final network than plain \gls{lr}: allocation reuses TESS matrix operations, while the enabled truncation further reduces memory and compute (Sec.~\ref{subsec:cost_learning}).

\subsection{Ablation Study}
\label{subsec:mechanism}

\begin{figure}[t]
    \centering
    \subfloat[Final accuracy and BWT.\label{fig:ablation-accuracy-bwt}]{%
        \includegraphics[width=\columnwidth]{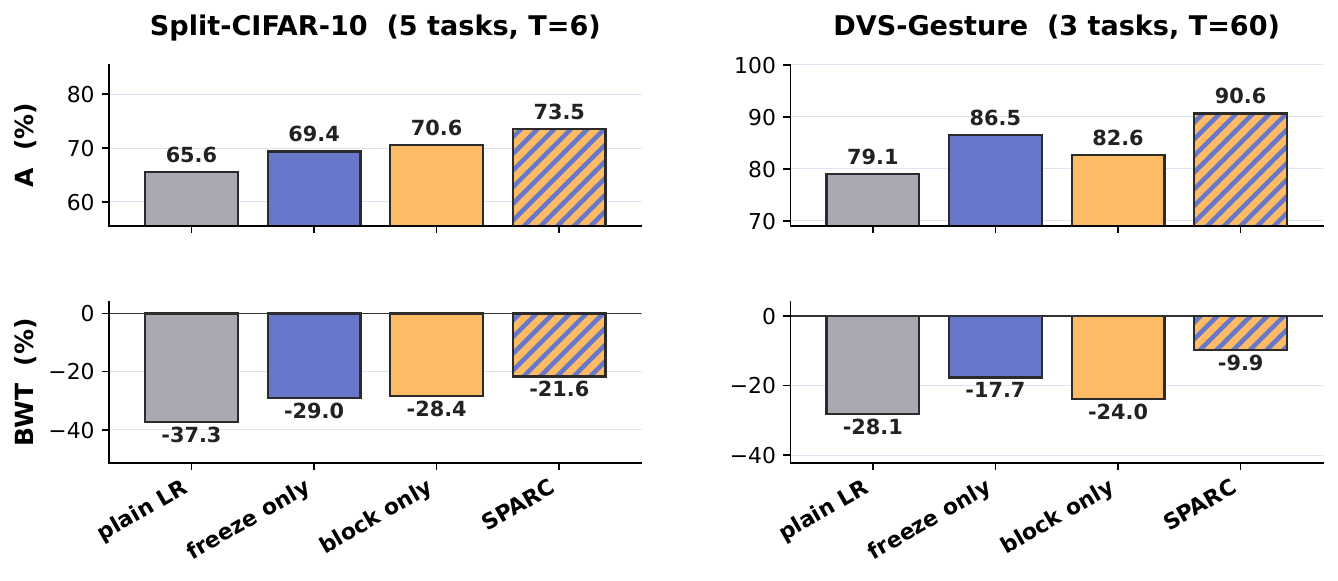}%
    }\\
    \subfloat[Forgetting across tasks.\label{fig:ablation-forgetting}]{%
        \includegraphics[width=\columnwidth]{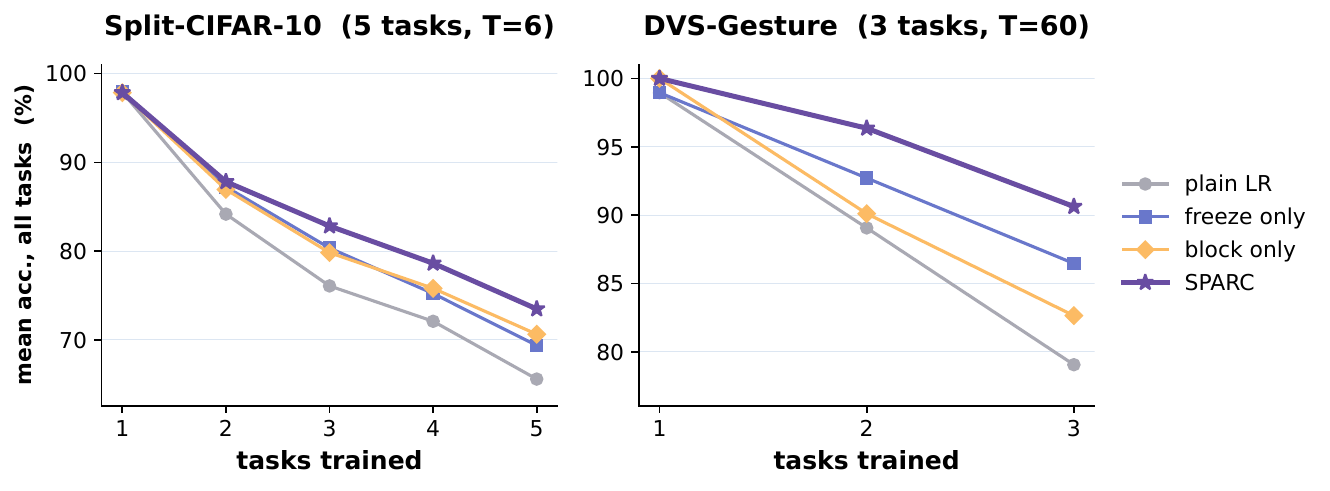}%
    }

    \caption{Mechanism ablation at the deployed Split-CIFAR-10 and
    DVS-Gesture settings.
    (a) Combining freezing with aligned block-sparse feedback gives the
    strongest final accuracy and BWT.
    (b) Mean accuracy over tasks seen; \tool's margin over plain LR appears
    at the second task and persists on both benchmarks.}
    \label{fig:ablation}
\end{figure}

Figure~\ref{fig:ablation-accuracy-bwt} isolates \tool's two mechanisms: freezing retains past knowledge, while aligned block-sparse feedback enables class-incremental, task-free readout at negligible structural cost. Each improves plain \gls{lr}, but their combination performs best on Split-CIFAR-10 and DVS-Gesture; neither component dominates. Figure~\ref{fig:ablation-forgetting} confirms that this is not a final-checkpoint artifact: \tool separates from plain LR when the second task arrives and maintains the gap throughout both sequences.

\subsection{The Cost of Learning}
\label{subsec:cost_learning}

We directly test literature claims that on-device learning requires a local alternative to \gls{bptt}~\cite{reckon, neurotrain, spikerll}, comparing \gls{tess} (used by \tool and plain LR) with \gls{bptt}.

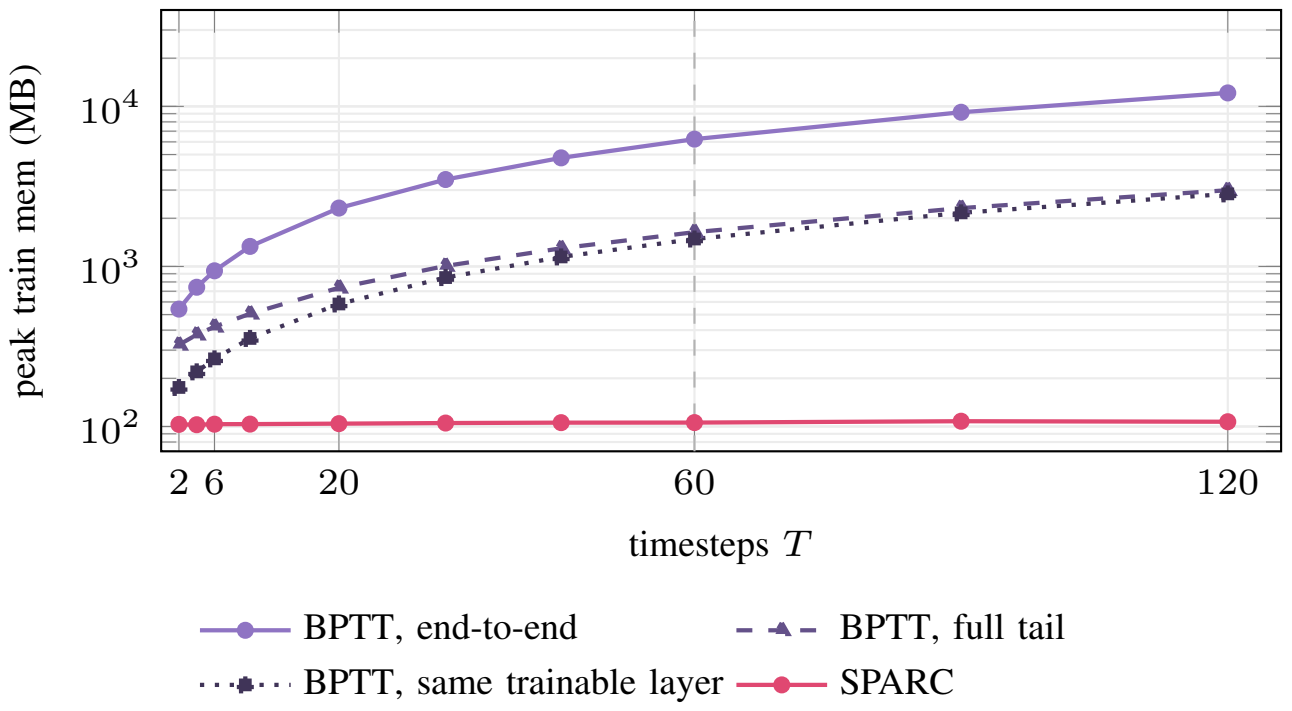
\begin{figure}[t]
\centering
\definecolor{toolpink}{RGB}{224,72,113}
\definecolor{toolblue}{RGB}{68,102,169}
\definecolor{toolpurple}{RGB}{143,116,195}
\begin{tikzpicture}
\begin{axis}[
    width=\linewidth, height=4.5cm,
    xlabel={timesteps $T$}, ylabel={peak train mem (MB)},
    ymode=log, log basis y=10,
    xmin=0, xmax=126, ymin=70, ymax=40000,
    xtick={2,6,20,60,120},
    legend to name=memorylegend,
    legend columns=2,
    legend style={font=\scriptsize, draw=none, fill=none},
    legend cell align=left,
    tick label style={font=\scriptsize},
    label style={font=\scriptsize},
    grid=both, grid style={gray!15},
]
\addplot[thick, mark=*, mark size=1.2pt, toolpurple] coordinates
    {(2,541.4)(4,741.4)(6,939.1)(10,1333.9)(20,2316.1)(32,3491.6)(45,4767.9)(60,6241.6)(90,9189.2)(120,12134.1)};
\addlegendentry{\gls{bptt}, end-to-end}
\addplot[thick, dashed, mark=triangle*, mark size=1.4pt, toolpurple!70!black] coordinates
    {(2,323.8)(4,374.7)(6,419.9)(10,506.5)(20,736.0)(32,1006.9)(45,1298.3)(60,1637.0)(90,2314.6)(120,2994.1)};
\addlegendentry{\gls{bptt}, full tail}
\addplot[thick, dotted, mark=square*, mark size=1.2pt, toolpurple!45!black] coordinates
    {(2,175.4)(4,219.5)(6,264.6)(10,355.0)(20,582.8)(32,852.2)(45,1146.4)(60,1483.9)(90,2161.5)(120,2840.0)};
\addlegendentry{\gls{bptt}, same trainable layer}
\addplot[thick, mark=*, mark size=1.2pt, toolpink] coordinates
    {(2,103.1)(4,102.6)(6,103.4)(10,103.5)(20,104.2)(32,105.1)(45,105.7)(60,105.8)(90,107.9)(120,107.1)};
\addlegendentry{\tool}
\draw[gray!60, dashed] (axis cs:60,70) -- (axis cs:60,40000);
\end{axis}
\end{tikzpicture}
\\[2pt]
\ref{memorylegend}
\caption{Peak training memory versus $T$ at the deployed DVS-Gesture configuration (split~4). \tool is constant in $T$; all three \gls{bptt} scopes are linear in it. The dashed rule marks the deployed $T{=}60$.}
\label{fig:memory}
\end{figure}

\paragraph{The learning-cost ledger}
Figure~\ref{fig:memory} shows that local-rule peak training memory is constant in $T$ whereas every \gls{bptt} arm grows linearly with it. At the deployed DVS-Gesture point ($T{=}60$) \tool needs $0.11$~GB against $1.64$~GB for full-tail \gls{lr} \gls{bptt} ($\approx15.5\times$), and $6.24$~GB for end-to-end \gls{bptt} on the same backbone ($\approx59\times$), which is the regime of the from-scratch baselines in Table~\ref{tab:soa}. The saving comes from replacing \gls{bptt} with \gls{tess} and from the truncation that locality permits. As expected the local rule advantage is modest at static-image timestep counts \cite{ottt, sltt} and decisive for long event sequences ($2.3\times$ at $T{=}6$ on Split-CIFAR-10, rising to $15.5\times$ at $T{=}60$).

All figures are peak allocated GPU memory over one training update, at each benchmark's own training batch size, following the measurement protocol of prior work~\cite{sltt, ottt}. This is a conservative estimate of deployed cost, since fixed framework overhead weighs more heavily on \tool's smaller footprint than on \gls{bptt}'s larger one.

\paragraph{One trainable layer}
\label{subsec:truncation}
TESS gives every layer a local readout at no extra architectural cost and makes its update independent of later layers; \tool can therefore truncate the tail after its first trainable layer without retraining. This changes the insertion-point preference: \gls{bptt} favors later cuts that retrain fewer layers, whereas \tool favors earlier cuts where its readout peaks.

At the deployed DVS-Gesture split~4, truncation reduces writable weights from $8.26$~M to $1.18$~M and leaves only one readout to evaluate. Figure~\ref{fig:insertion-accuracy-bwt} shows no accuracy penalty: generally this readout matches or exceeds plain \gls{lr}'s oracle over the untruncated tail.

Across the full sweep (Fig.~\ref{fig:insertion-cost}), \tool's training memory and synaptic-operation count never exceed full-tail plain \gls{lr} under \gls{tess} at any cut, and coincide with it exactly at split~7, where there is nothing left to truncate. Full-tail \gls{tess} in turn stays below \gls{bptt} at every cut on both datasets. \tool's cost tracks the trained layer's own dimensions rather than falling monotonically with depth. Across methods and insertion points, the common replay buffer remains one to two orders of magnitude below full-tail local-rule memory and further below \gls{bptt}; the trainable tail, not replay, dominates memory pressure.

\paragraph{Synaptic operations}
Figure~\ref{fig:insertion-cost} shows the same operation-count advantage across insertion points, until split~7, where the tail collapses to one convolution and \gls{bptt} becomes cheaper. This does not contradict \gls{bptt}'s hardware infeasibility, which comes from the type of operation, a global backward pass with transposed access, rather than the count.

\paragraph{On \gls{bptt} memory usage}
Temporal reversibility~\cite{trevsnn} and gradient checkpointing with lossless spike compression~\cite{snnckpt} shrink \gls{bptt}'s memory further, trading it for recomputation. Neither makes it constant in $T$, and neither removes the global backward pass, the transposed weight access, or the need to buffer a sequence and traverse it in reverse before an update is emitted, the properties that keep \gls{bptt} off a local, event-driven accelerator, and which a local rule avoids by construction.

\subsection{Comparison with SOA Methods}
\label{subsec:comparison}

\begin{table}[t]
\centering
\caption{Published \gls{snn} class-incremental results}
\label{tab:soa}
\small
\setlength{\tabcolsep}{3pt}
\begin{tabular}{llrrrr}
\toprule
Method & Backbone & $T$ & Retained & $A$ (\%) & BWT \\
\midrule
\multicolumn{6}{l}{\textit{Split-CIFAR-10, $5\times2$ classes}} \\
CATFormer~\cite{catformer}       & SpikFormer  & n.s. & $64.2$~kB/task & 89.3 & n.s. \\
\textbf{SPARC (ours)}            & VGG-9       & 6    & 1000 / 5.9     & \textbf{73.5} & $-21.6$ \\
\textbf{SPARC (500)}             & VGG-9       & 6    & 500 / 2.93     & \textbf{68.8} & $-29.2$ \\
STAER~\cite{staer}               & ResNet-19   & 4    & 500 / 1.47     & 65.7 & $-29.2$ \\
DER++~\cite{staer}               & ResNet-19   & 4    & 500 / 1.47     & 60.8 & $-31.4$ \\
ER~\cite{staer}                  & ResNet-19   & 4    & 500 / 1.47     & 50.8 & $-53.2$ \\
\textbf{SPARC (200)}             & VGG-9       & 6    & 200 / 1.17     & \textbf{60.0} & $-42.7$ \\
STAER~\cite{staer}               & ResNet-19   & 4    & 200 / 0.59     & 51.2 & $-35.5$ \\
DER++~\cite{staer}               & ResNet-19   & 4    & 200 / 0.59     & 51.1 & $-32.7$ \\
ER~\cite{staer}                  & ResNet-19   & 4    & 200 / 0.59     & 39.0 & $-70.3$ \\
Energy-aware~\cite{energyaware}  & LIF convnet & 50   & n.s.           & 61.3 & $-60.2$ \\
\midrule
\multicolumn{6}{l}{\textit{DVS-Gesture, $4{+}4{+}3$ classes, $T{=}60$}} \\
Energy-aware~\cite{energyaware}  & LIF convnet & 60   & n.s.           & 91.9 & $-5.9$ \\
\quad (replay only)              & LIF convnet & 60   & n.s.           & 74.5 & $-25.4$ \\
\textbf{SPARC (ours)}            & VGG-9       & 60   & 192 / 22.7     & \textbf{90.6} & $-9.9$ \\
\midrule
\multicolumn{6}{l}{\textit{CIFAR10-DVS, $5\times2$ classes}} \\
CATFormer~\cite{catformer}       & SpikFormer  & n.s. & n.s.           & 87.1 & n.s. \\
\textbf{SPARC (ours)}            & VGG-9       & 10   & 180 / 3.96     & \textbf{42.4} & $-36.9$ \\
Energy-aware~\cite{energyaware}  & LIF convnet & 50   & n.s.           & 49.7 & $-25.1$ \\
\bottomrule
\end{tabular}
\par\vspace{2pt}
\parbox{\columnwidth}{\footnotesize Retained: samples / size in MiB unless stated. SPARC uses local learning; the other methods use BPTT. n.s.: not stated. BWT is negative for forgetting.}
\end{table}

Table~\ref{tab:soa} places these results against the field. Notably most of the methods considered here are trained globally while \tool{} adapts a single layer through a local rule. On DVS-Gesture, \tool trails Energy-Aware Spike Budgeting~\cite{energyaware} by $1.3$~pp under its $4{+}4{+}3$, $T{=}60$ protocol ($90.6$ vs $91.9$), while Energy-Aware trains its whole network end to end with \gls{bptt}. As a reference, full-training of the VGG9 net used by us would be $6.2$~GB vs $0.11$ ($59\times$ less). On Split-CIFAR-10, \tool accuracy leads by $12.2$~pp ($73.5$ vs $61.3$) with roughly $2.8\times$ less forgetting. At matched Split-CIFAR-10 buffers, \tool improves STAER/snn-DER++/snn-ER~\cite{staer} accuracy: $68.8$ vs $65.7/60.8/50.8$ with $500$ samples and $60.0$ vs $51.2/51.1/39.0$ with $200$. CATFormer~\cite{catformer} leads outright ($89.3$) using in-domain first-task features, per-task threshold fitting by \gls{bptt}, and two-pass gated inference; \tool uses an out-of-distribution backbone and one task-free pass, a different scope and complexity.

Additionally we limit-tested \tool on Split-CIFAR-100 (10 tasks x 10 classes). Comparing with the only other local continual learning technique demonstrated on this dataset AGMP~\cite{agmp}, \tool gains $4.9$~pp accuracy ($36.3$ vs $31.4$) while training one layer rather than a full online ResNet-18. This is nevertheless a limit application for local methods and it is where more complex methods lead (CATFormer~\cite{catformer} $68.3$, DSD-SNN~\cite{dsdsnn} $60.5$).

\subsection{Backbone Analysis}
\label{subsec:backbone}

Figure~\ref{fig:donor-transfer} reports accuracy across pretraining donors. We observe that DVS-Gesture is markedly less sensitive to the choice of donor than Split-CIFAR-10. The donor sweep at different insertion layers shows that an earlier split narrows donor sensitivity on both datasets, yet at the same split~4, cross-modality donors trail the in-domain donor by $33.2$~pp on Split-CIFAR-10 (versus $52.3$~pp at split~6) against just $5.2$~pp on DVS-Gesture.

\section{Conclusion}
\label{sec:conclusion}

We presented \tool, a task-free \gls{cil} method combining one-bit \gls{lr} with local \gls{tess} adaptation in deep \glspl{snn}.  \tool improves over conventional \gls{lr} while maintaining constant peak training memory with sequence length, unlike \gls{bptt}.
Despite these results, limitations remain. Accuracy is modest on very large datasets such as Split-CIFAR-100, as common for local training methods. \tool assumes known stage boundaries and preconfigured class/channel capacities, and currently partitions only convolutional channels. Finally, deployment costs remain unmeasured because suitable neuromorphic hardware is not yet available; moreover, differences in backbones, pretraining, and protocols across published results preclude uniform state-of-the-art comparisons.

\bibliographystyle{IEEEtran}
\bibliography{bibliography}

\end{document}